# Integrating Soft Robotics with ROS: A Hybrid Pick and Place Arm*

Ross M. McKenzie, Thomas W. Barraclough, and Adam A. Stokes, *Member, IEEE*

*Abstract*—Soft robotic systems present a variety of new opportunities for solving complex problems. The use of soft robotic grippers, for example, can simplify the complexity in tasks such as the of grasping irregular and delicate objects. Adoption of soft robotics by academia and industry, however, has been slow and this is, in-part, due to the amount of hardware and software that must be developed from scratch for each use of soft system components. In this paper we detail the design, fabrication and validation of an open-source framework that we designed to lower the barrier to entry for integrating soft robotic subsystems. This framework is built on ROS (the Robot Operating System) and we use it to demonstrate a modular, soft-hard hybrid system which is capable of completing pick and place tasks. By lowering this barrier to entry we hope that system designers and researchers will find it easy to integrate soft components into their existing ROS-enabled robotic systems.

## I. INTRODUCTION

### A. The Barrier to Entry for Soft Robotics

The lack of open source hardware and software for soft robotics creates a significant barrier to entry for researchers who wish to conduct research into soft robotic systems. In this paper we detail our development of an open source, modular, soft-hard hybrid robot whose components can be easily manufactured and integrated into existing or robotic systems. By lowering this barrier to entry we aim to provide the informatics community with a new opportunity to explore the benefits of soft robotics for a wide range of tasks.

A robotic system contains a combination of hardware, sensing, and control. The established control paradigms for hard robot systems are based on assumptions that the links in an assembly are rigid, and that the joint angles can be measured using encoders. Soft robotic components do not have rigid links and so the established control methods cannot be applied [1]. Soft robotics researchers require control-hardware and control-software which is designed for their specific actuation and sensing techniques.

The advantages of integrating soft robotic components with other types of robotic systems is demonstrated by their use in haptic classification (used in combination with a Baxter robot) [2] and in object retrieval (used in combination with a Roomba wheeled-hard robot) [3]. These two examples are hybrids, and share many common system components, but the system designers used their own bespoke control systems. This repetition of existing work slows the rate of soft robotic research and reduces innovation. There are some resources available to researchers when they are designing and fabricating the soft components themselves. F. Ilievski, et al. [4] detail the fabrication of soft robotic actuators, pneunets. Pneunets are pneumatic-channels which are molded into the interior of a soft robot and they actuate by expanding due to pressurized gas. Instructions for building the specific soft grippers we use in this paper are presented in a step-by-step format by B. Finio, et al. [5]. This existing body of work means that researchers without a materials science background have enough information to fabricate soft robotic grippers.

Another significant resource is the soft robotic toolkit [6]. This website holds a repository of information on soft robotics and also includes an open source design for a manual pneumatic control board. This board contains a microcontroller in order to enact human inputs (from potentiometers), and therefore could be adapted for automatic control by adding custom code. However, it uses a number of bread boards with wiring in between, and is therefore large. Our design of a compact circuit board with an included program for computer control will be easier to integrate into other robotic designs.

The resources for building a full robotic system which integrates hard and soft components are much more limited.

The open source hardware and software which we present in this paper provide a new resource for research into integrated soft robotic systems and will lower the barrier to entry for researchers. We provide easy to fabricate modules for: 1.) a hard robotic arm; 2.) a soft robotic gripper, and 3.) the electronics and software used to build a fully controlled system.

### B. Soft Robotics for Gripping and Manipulation Tasks

Robotic grasping can use a number of methods that fall into two main groups: 1.) fingertip grasps, when an object is pinched between contact points and 2.) enveloping grasps, where the fingers and palm of a gripper wrap around an object [7]. Enveloping grasps are better at restraining objects but can be difficult to achieve with hard robotics unless the shape and size of the target object are known before the gripper is developed. By making grippers soft an enveloping grasp becomes much easier as the gripper can conform to the shape of an object without external control [4]. These soft grippers can also handle more delicate objects, such as fruit [8] [9] [10] and biological samples [11].

There are numerous designs and actuation methods for soft grippers, such as those that use the jamming properties of granular materials under a vacuum [12], electro-adhesion [13] and shape memory alloys [14].

We used the four fingered pneumatic soft gripper as detailed in B. Finio, et al. (2013) [5]. We chose this gripper because the fabrication process is well documented. To

*Research supported by EPSRC.

R. M. McKenzie, T. W. Barraclough, and A. A. Stokes are with the Engineering Department, University of Edinburgh (phone: 0044-131-6501000; fax: 0044-131-6507475; e-mail: R.M.Mckenzie@ed.ac.uk, T.Barraclough@ed.ac.uk, Adam.Stokes@ed.ac.uk).

simplify the control system we used a single pneumatic control line, rather than having a separate channel in each finger.

*C. The Motivation for Using Hybrid Hard and Soft Systems*

By using a hybrid hard-soft system we allow the robot to use the advantages of both hard and soft robotics. This idea was demonstrated in our previous work, A. A. Stokes, et al. [3], where we showed that the advantages of speed and accuracy offered by hard robotics could be combined with the versatility of soft robotics. We use a similar design strategy here, for a pick and place task. For this grasping and manipulation task precise and fast positioning of the end effector is best achieved by a hard robotic arm, while a compliant, enveloping grip is most easily achieved using a soft gripper.

*D. Integrating Pneumatic Soft Grippers with The Robot Operating System (ROS)*

Standardization is an important step in fostering innovation. An example of this benefit of standardization is found in the standardized control language (G-Code) used in Computer Numerical Control (CNC) machines [15]. Before this standardization each manufacturer had used their own language and it was difficult for companies to change machines making innovation difficult to propagate. Similarly, innovation in soft robotics is currently held back by the need to implement a bespoke control system for each new robotic system.

Robotics is moving towards standardization and commercially available robots such as the Baxter [16] and PR2 [17] are designed to work with a standardized software system called the Robotic Operating System(ROS) [18]. ROS is a framework that allows for easy integration of ROS packages which have different functionalities. This ease of integration is enabled by using independent nodes to run programs whilst enabling communication through message carriers called topics. These nodes and topics have standardized protocols allowing easy use of other researcher's individual nodes and topics or structures of multiple nodes and topics.

This standardized system encourages collaborative development by allowing different researchers to quickly find and use the work of others which has been published in the global ROS package repository. This collaborative approach avoids the need for researchers to repeat the work of others when creating their new research platform. We chose ROS as our system controller to make it easier to integrate our whole system, or its modular components, with other robots—for example a wheeled platform.

*E. Design of a Modular, Open Source, Hybrid System for Pick and Place Tasks*

Pick and place tasks are found in a variety of industrial processes. Many pick and place tasks such as those in agriculture or ecommerce order fulfillment have not been automated due to the requirement for delicate or compliant grasping [19] and the variety in size, shape and rigidity of the target objects involved [20]. By designing this open source package around a pick and place task we can demonstrate integration of both hard and soft modules into a system which is capable of performing task-oriented work. Fig. 1. Provides an overview of the system.

*1) Design of The Control Software*

*a) Design of the ROS Control*

We designed our software to be modular; the ROS nodes execute specific tasks and it is easy to exchange them for new nodes. The arm control is, therefore, completely separate from the soft robotic control. We spanned the hard and soft control sections of the code with an overall system controller node which sends instructions to the arm and to the pneumatic controllers. The system controller also monitors the hard and soft controllers so that it can wait for an instruction to be completed before issuing another. Fig. 2. Shows the layout of the ROS nodes and topics.

*b) 3-Axis Control*

There are pre-existing ROS packages for motion planning with robot arms, such as MoveIt! [21], however these packages are designed to control complex research-arms with more than three degrees of freedom. For our simple rigid link robot arm we designed a new program which used analytical solutions for the joint angles.

*c) PID control*

The pressure control node runs a PID loop using the values published by the pressure sensor. This design allows initial actuation to the component and it maintains pressure over time. We designed the controller to maintain pressure between an upper and lower boundary to avoid rapid oscillations in the applied pressure.

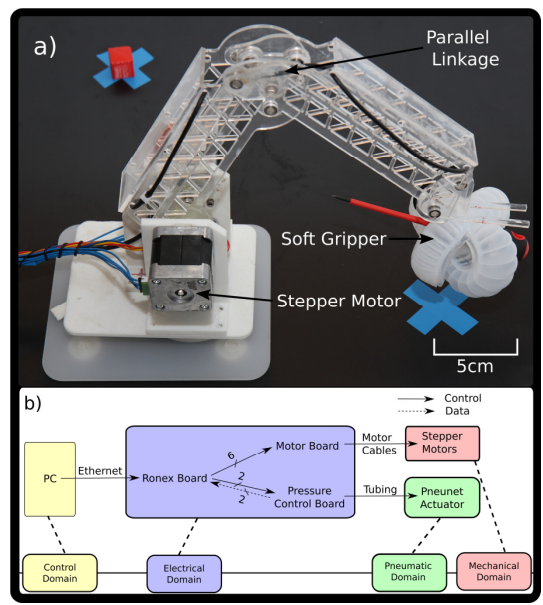

Figure 1. An overview of the open source system. A) A picture of the soft-hard hybrid arm. B) A Block diagram of the system

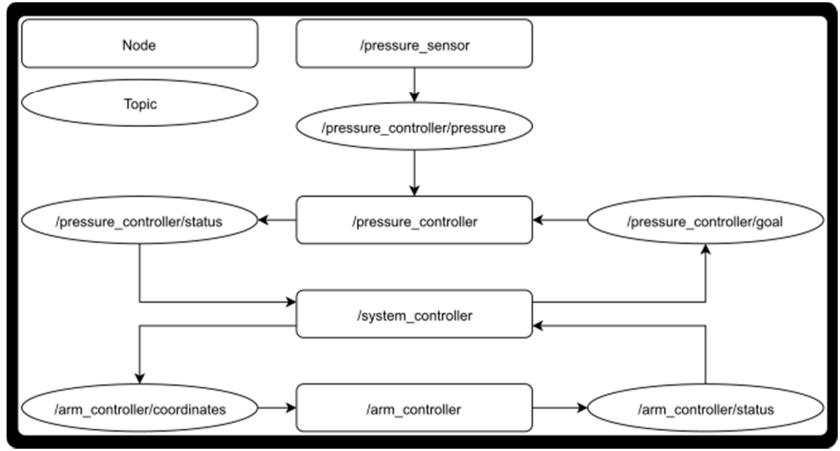

Figure 2. A graph showing the organization of ROS nodes and topics used to control the system. The soft robotic control is in the top half while the bottom half is the arm control.

*2) Design of the Hardware*

*a) The Design of the Embedded Electropneumatic Control System*

To allow direct control from a PC we used a Ronex Board from Shadow robotics [22]. The Ronex board provides two 5V power sources and 12 5V Digital Input/Output (IO) pins. The board is fully ROS enabled and so our ROS nodes could easily control the electropneumatic system.

Our system also comprised a motor board and a pneumatic control board. The motor board is designed to run 3 stepper motors using 6 IO pins while the pneumatic control board uses 4 IO pins to run a pump, valve and pressure sensor.

*b) Design of the Rigid Link Robot Arm*

We based the arm on an existing open source arm design, the Dobot Arm [23], which we altered to be lightweight and easier to manufacture using our available tools. The arm involves a parallel linkage so that the end effector is always parallel to the surface upon which the arm is positioned. This feature removes some of the complexity in controlling the arm.

Our arm does not require the precision of industrial robotic arms as the wide, enveloping grasp of our mounted soft gripper allows for the robust grasping of objects that are not directly beneath the end effector. Our system can accomplish manipulation tasks reliably and only costs around $100, including the pneumatic system, but excluding the Ronex board. Conventional research arms for similar tasks, for example a Kuka YouBot arm [24], can cost around $18,000 and hobbyist arms, such as the official Dobot arm, will cost $900 [23]. These arms are designed to have micrometer precision, as conventional robotic grippers can only tolerate a low error rate when picking up small or complex objects. By removing this requirement our arm can be made cheaper by an order of magnitude, as it does not need highly accurate motors, encoders, and force sensors on the joints and can be made from low cost plastic.

*c) Design of the Soft Robotic Gripper*

We used the design for a soft gripper detailed in B. Finio, et al. [5]

*F. Design of Validation Experiments*

The arm must operate below a reasonable level of position error, not stall its motors and not accumulate position error over time through loss of steps. We tested these three factors by instructing the arm to make six 50 cm movements within its plane and then we measured its position. Analysis of these results will enable us to find the positioning error in these movements. After each motion we instructed the arm to return to its exact starting location and we measured the difference between the start and end locations to obtain another error value. This error allows us to determine if steps are being lost by the motors, because the arm should return to the same starting position. We calculated the mean-error for position and step-loss in the horizontal and vertical directions. The systematic error in the measurements is 0.05 mm from the Vernier caliper.

We performed a similar test with rotation by instructing the arm to turn 90 degrees and then return to its starting position, we measured the error after each motion and three repeats were used. The systematic error in this measurement is 0.5 degrees arising from the protractor.

TABLE I. THE RESULTS OF THE VALIDATION TEST

|  | Horizontal (mm) | Vertical (mm) | Perpendicular to Arm Plane (mm) |
|---|---|---|---|
| Mean Error After Moving 50cm | 4.0±0.9 | 2.5±0.4 | N/A |
| Mean Error After Return | 0.50±0.08 | 0.26±0.06 | N/A |
| End Effector Backlash | 3.5±0.5 | 1.5±0.5 | 10.0±0.5 |

We also performed a test on end effector backlash, i.e. how far the end effector could be moved without forcing the motors. We performed this test by moving the end effector by hand in three perpendicular directions and we measured the distance between extreme positions. We estimated the measurement error in this method to be 0.5 mm.

*G. Design of the Soft Grasping and Manipulation Task*

To test the ability of the arm to grasp and manipulate objects we placed a target box and a target object on a table. We chose the position so that the height of the end effector and rotation and extension of the arm would all need to change in order for the task to succeed. To test the delicacy of the soft gripping we chose one object, a tomato. We chose a second object, a syringe, as it had an asymmetric shape. These two shapes allowed us to test the ability of the gripper to pick up objects using the same control input. The only change made to the instructions between the two tests was to place the end effector 1 cm lower during the syringe grasp to account for the difference in height

.

## II. RESULTS AND DISCUSSION

*A. System Fabrication and Verification*

We fabricated the arm using a combination of 3D printed and laser cut components. We fabricated the soft gripper using Ecoflex 00-50 and a 3D printed mold. Further assembly information is included in the experimental section.

*B. Evaluation of Arm Performance*

Table 1 shows the results of the validation tests. The arm is precise enough to provide a platform for a soft gripper. Its end effector, however, can be moved perpendicular to the plane of the arm by 1 cm. The backlash in this direction could mean that it is not guaranteed to perform well in soft grasping tasks with small (2-3 cm across) grippers. This result is mitigated as we measured from extreme to extreme and so the true effect on precision is only 0.5 cm. The end effector also returns to the same rest position when released and so letting the arm settle after each movement removes the effect of the backlash. This backlash error arises from non-perfect fits between parts of the arm as well as deformation in the material of the arm.

Table 1 shows the results of our tests to assess increasing inaccuracy over time due to step loss. Using the starting height of the position error tests and assuming that the second joint was vertically downwards we calculated the joint angles to be 65 degrees for the first and 90 degrees for the second joint. The change of one step (1.8 degrees) in either direction from either joint will produce a minimum of 4.3 mm positioning error. This result demonstrates that our arm is not losing steps. The error is caused by our use of microstepping [25], which is a method used to give stepper motors a higher resolution than their step size. Microstepping can cause hysteresis which is why our motors did not return to their exact starting location. This effect does not cause accumulation of position error over time.

*C. Results of The Pick and Place Task*

We instructed the arm to execute the two manipulation tasks and it functioned correctly and within an acceptable level of precision. This is demonstrated by a video that we have submitted along with the paper at http://ieeexplore.ieee.org and also as two separate videos at:

http://edin.ac/2cKLWKS

http://edin.ac/2ccasze

These videos demonstrate the ability of our system to pick up both regularly and irregularly shaped objects with the same control input to the simple soft gripper. Fig. 3. Shows snapshots from the tomato manipulation test.

*D. Scope for Development*

*1) Improvements to Arm Accuracy*

As the arm was only losing one step after an average of two motions the loss of steps is unlikely to be the cause of the imprecision. The inverse kinematics were checked by hand and are accurate. Therefore, the main cause of inaccuracy in the arm must have been caused by the inaccuracy in its starting location being propagated through the inverse kinematics. Introducing sensors to the arm, such as end stops or hall effect sensors would provide a way to determine an accurate starting position at every restart.

Gearboxes could further reduce the chance of the arm stalling by increasing torque but would reduce travel speed.

*2) Additional Modules for The ROS Packages*

We designed the open source package to be modular and as such more nodes can be added to it. These could include curvature sensors such as were demonstrated by B. S. Homburg, et al. [2] or nodes to control other actuation methods such as a vacuum pump to run a jamming gripper such as the one demonstrated by J. R. Amend, et al. [12].

## III. CONCLUSIONS

The lack of open source resources for those who wish to conduct research with soft robotics impedes progress in two areas: 1.) engineering work on soft robotics is impeded due to a lack of standardization and the time spent recreating the work done by others, and 2.) the need to create bespoke hardware provides a significant barrier to entry for those whose primary research field is informatics rather than engineering.

Our work provides the first open-source computer control system specifically developed for soft robotics that includes hardware and software. Additionally, we have designed, fabricated, and demonstrated the first open-source soft-hard hybrid robot for gripping and manipulation tasks.

By creating this resource for soft robotics we aim to allow informatics researchers to integrate soft robotic systems with existing robots and to use soft robotics without needing to design hardware. This step towards standardization should allow for further research into the control of soft robotics and the use of soft robotics for complex tasks.

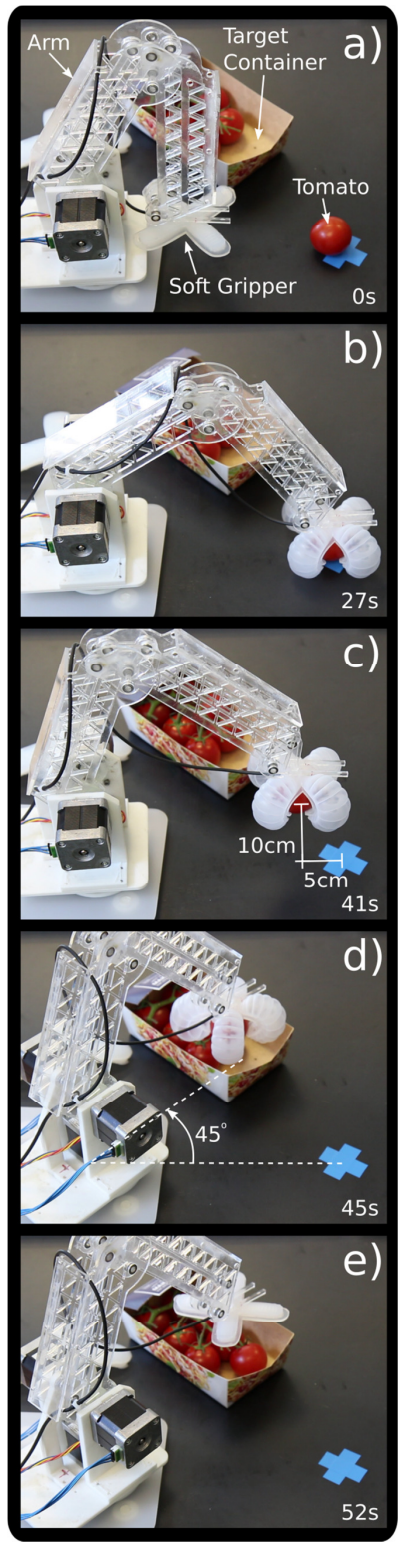

Figure 3. Still frames from a video of the arm performing a manipulation task with a tomato. In this task a tomato is grasped, manipulated and placed in a box

Soft robotics researchers will also benefit from our work as they have the option of a standardized controller which will integrate with other robotics platforms, and therefore they do not need to recreate all the hardware and software from scratch.

Our platform need not only serve an academic purpose as we have used it to demonstrate the ability of a simple hybrid system to complete pick and place tasks. This work could provide a basis for the industrial development of soft-hard hybrid automation solutions.

IV. EXPERIMENTAL

A. Hardware

*1) 3D Printing*

All 3D printed components were made from 1.75 mm, white PLA purchased from PrintMe3D for £18 and were printed on a Wanhou Duplicator 4X

*2) Arm*

A full 3D schematic for the arm can be found at http://www.homepages.ed.ac.uk/astokes2/research_BSR.html#ROS. A diagram of the arm detailing the components is available in the supplementary material at http://ieeexplore.ieee.org, it is also available at http://www.homepages.ed.ac.uk/astokes2/research_BSR.html#ROS

We purchased the three stepper motors from RS Components #829-3500 for £31.94 each. The motor holders, arm plate and linkage 0 were 3D printed and the remaining components were laser cut from 1 $m^2$ acrylic purchased from Acrylic Online for £17 on an Epilog Laser Fusion. To start the arm all of the 9mm holes had 9 mm - 5 mm bearings affixed in them with superglue. Then the parallel linkage was assembled by attaching the end effector to linkages 3 to 0 in series. The minor arm was then assembled around the linkage. We then attached the major arm sections. We then assembled the base, mounted the stepper motors and attached the arm.

*3) Motor Control Board*

The motor control board schematics can be found at http://www.homepages.ed.ac.uk/astokes2/research_BSR.html#ROS. We purchased the three stepper motor drivers from MiniInTheBox.com part number #04550672 for $4.99 each. We milled the motor control board using an LPKF ProtoMat S63. The drivers were soldered to the board by hand.

*4) Pressure Controller*

The pressure controller board schematics are available at http://www.homepages.ed.ac.uk/astokes2/research_BSR.html#ROS. The pump and valve used in the pressure controller were taken from a dismantled digital blood pressure meter. The MOSFET was purchased from RS components #541-0610 for £0.92. The resistors were part of a pack purchased from RS Components #156-2565 for £36.03. The pressure sensor was purchased from DigiKey Electronics #480-6591-ND £34.89 and the transistor was purchased from DigiKey Electronics #2SC3940ASACT-ND for £0.32. We milled the pressure controller board using an LPKF ProtoMat S63. The components were soldered to the board by hand.

*5) Soft Robotic Gripper*

We fabricated the soft robotic grippers using EcoFlex 00-50 purchased from Bentley Advanced Materials for £175.56 and a 3D printed moulds, a full methodology for the fabrication of the grippers can be found in B. Finio, et al. [5].

*B. Software*

The full software package used to run the robot in this paper is available open-source at http://www.homepages.ed.ac.uk/astokes2/research_BSR.html#ROS.

*1) Arm Control*

The arm controller uses analytical inverse kinematic solutions to determine required joint angles to reach a Cartesian coordinate published in the arm control topic. The program then finds the nearest space possible with the discrete steps of each motor and updates the joint angles and Cartesian position with their true values.

*2) Pressure Control*

The pressure sensor node uses two I/O wires to communicate with an $I^2C$ sensor then publishes the pressure reading on a pressure topic. The pressure control node runs a PID loop using the values published in the pressure topic. As the Ronex I/O cables are digital we used Pulse Width Modulation(PWM) [26] to provide variable power to the pump and valve. The I-term was halved whenever the pressure passed the optimum and its maximum value was limited.

*3) ROS system control*

The system controller reads a G-Code file and queues the instructions, it uses G01 for both movement and grasping commands. The controller then publishes each instruction on either the pressure control or arm control topic. The system controller subscribes to the arm status and pressure status topics to determine when to start the next instruction

. ACKNOWLEDGMENT

This study was supported by EPSRC via the Robotarium Capital Equipment and CDT Capital Equipment Grants (EP/L016834/1) and the CDT in Robotics and Autonomous Systems

V. REFERENCES

[1] D. Ross, M. P. Nemitz and A. A. Stokes, "Controlling and Simulating Soft Robotic Systems: Insights from a Thermodynamic Perspective," *Soft Robotics,* 2016.

[2] B. Homberg, R. Katzschmann, M. Dogar and D. Rus, "Haptic Identification of Objects using a Modular Soft Robotic Gripper," in *IEEE/RSJ International Conference on Intelligent Robots and Systems (IROS)*, 2015.

[3] A. Stokes, R. Shepherd, S. Morin, F. Ilievski and G. Whitesides, "A Hybrid Combining Hard and Soft Robots," *Soft Robotics,* vol. 1, no. 1, pp. 70-74, 2014.

[4] F. Ilievski, A. D. Mazzeo, R. F. Shepherd, X. Chen and G. M. Whitesides, "Soft Robotics for Chemists," *Angewandte Chemie International Edition,* vol. 50, no. 8, pp. 1890-1895, 2011.

[5] B. Finio, R. Shepherd and H. Lipson, "Air-Powered Soft Robots for K-12 Classrooms," in *3rd IEEE Integrated STEM Education Conference*, 2013.

[6] "The Soft Robotics Toolkit," [Online]. Available: http://softroboticstoolkit.com/. [Accessed 01 09 2016].

[7] A. Bicchi and V. Kumar, "Robotic Grasping and Contact: A Review," in *IEEE International Conference on Robotics and Automation 1*, 2000.

[8] J. D. Tedford, Developments in robot grippers for soft fruit packing in New Zealand, vol. 8, Cambridge : University Press, 1990, p. 5.

[9] E. J. van Henten, J. Hemming, B. A. J. van Tuijl, J. G. Kornet, J. Meuleman, J. Bontsema and E. A. van Os, "An Autonomous Robot for Harvesting Cucumbers in Greenhouses," *Auton. Robots,* vol. 13, no. 3, pp. 241-258, November 2002.

[10] Soft Robotics Inc., "Homepage," [Online]. Available: http://www.softroboticsinc.com/. [Accessed 09 09 2016].

[11] G. K. C., B. K. P., P. Brennan, K. Jordan, L. Stephen, T. Dan, W. R. J. and G. D. F., "Soft Robotic Grippers for Biological Sampling on Deep Reefs," *Soft Robotics,* vol. 3, no. 1, pp. 23-33, 2016.

[12] J. R. Amend, E. Brown, N. Rodenberg, H. M. Jaeger and H. Lipson, "A positive pressure universal gripper based on the jamming of granular material," *IEEE Transactions on Robotics,* vol. 28, no. 2, 2012.

[13] J. Shintake, S. Rosset, B. Schubert, D. Floreano and H. Shea, "Versatile Soft Grippers with Intrinsic Electroadhesion Based on Multifunctional Polymer Actuators," *Adv. Mater.,* vol. 28, no. 2, pp. 231-238, January 2016.

[14] W. Wang, H. Rodrigue, H.-I. Kim, M.-W. Han and S.-H. Ahn, "Soft composite hinge actuator and application to compliant robotic gripper," *Composites Part B: Engineering ,* vol. 98, pp. 397-405, 2016.

[15] E. I. Association and A. N. S. Institute, Interchangeable variable block data format for positioning, contouring, and contouring/positioning numerically controlled machines, Washington, D.C.: The Department, 1980.

[16] Rethink Robotics, "Rethink Robotics - Baxter," [Online]. Available: http://www.rethinkrobotics.com/baxter/. [Accessed 05 09 16].

[17] Willow Garage, "Willow Garage - PR2," [Online]. Available: http://www.willowgarage.com/pages/pr2/overview. [Accessed 05 09 16].

[18] "ROS," [Online]. Available: http://www.ros.org/. [Accessed 05 09 2016].

[19] F. Rodríguez, J. C. Moreno, J. A. Sánchez and M. Berenguel, "Grasping in Agriculture: State-of-the-Art," in *Grasping in Robotics*, London, Springer-Verlag, 2013, pp. 385-409.

[20] C. Liang, K. Chee, Y. Zou, H. Zhu, A. Causo, S. Vidas, T. Teng, I. Chen and K. L. Cheah, "Automated Robot Picking System for E-Commerce Fulfillment Warehouse Application," in *The 14th IFToMM World Congress*, Taipei, 2015.

[21] I. A. Sucan and S. Chitta, "MoveIt!," [Online]. Available: http://moveit.ros.org/. [Accessed 06 09 16].

[22] Shadow Robotics, "Shadow Robotics - Ronex Board," [Online]. Available: http://www.shadowrobot.com/products/ronex/. [Accessed 06 09 2016].

[23] Shenzhen Yuejiang Technology Co. Ltd, "Dobot," [Online]. Available: http://dobot.cc. [Accessed 17 08 2016].

[24] Kuka, "youBot Price List," [Online]. Available: http://www.youbot-store.com/pricelist. [Accessed 09 09 2016].

[25] University of Texas, "Microstepping," [Online]. Available: http://users.ece.utexas.edu/~valvano/Datasheets/StepperMicrostep.pdf. [Accessed 08 09 2016].

[26] M. Barr, "Introduction to Pulse Width Modulation (PWM)," [Online]. Available: http://www.barrgroup.com/Embedded-Systems/How-To/PWM-Pulse-Width-Modulation. [Accessed 17 08 2016].